\documentclass[letterpaper, 10 pt, conference]{ieeeconf}  

\IEEEoverridecommandlockouts                              
\usepackage{amsmath} 
\usepackage{amssymb}  
\usepackage{verbatim}  
\usepackage{graphicx}
\usepackage{subfig}
\usepackage{array}

\usepackage{float}
\usepackage{placeins}
\usepackage{hhline}
\usepackage{multirow}

\graphicspath{imgs}

\title{\LARGE \bf
  Robust Registration and Geometry Estimation from Unstructured Facial Scans
}

\author{Maxim Bazik$^{1}$ and Daniel Crispell$^{2}$
  \thanks{$^{1}$M. Bazik is a Research Developer at Vision Systems Inc, Providence, RI 02903, USA
        {\tt\small max.bazik at visionsystemsinc.com}}%
\thanks{$^{2}$D. Crispell is a Principal Scientist at Vision Systems Inc, Providence, RI 02903, USA
        {\tt\small daniel.crispell at visionsystemsinc.com}}%
}

\begin{document}

\maketitle
\thispagestyle{empty}
\pagestyle{empty}

\begin{abstract}
Commercial off the shelf (COTS) 3D scanners are capable of generating point clouds covering visible portions of a face with sub-millimeter accuracy at close range, but lack the coverage and specialized anatomic registration provided by more expensive 3D facial scanners.
We demonstrate an effective pipeline for joint alignment of multiple unstructured 3D point clouds and registration to a parameterized 3D model which represents shape variation of the human head.
Most algorithms separate the problems of pose estimation and mesh warping, however we propose a new iterative method where these steps are interwoven.
Error decreases with each iteration, showing the proposed approach is effective in improving geometry and alignment.
The approach described is used to align the NDOff-2007 dataset, which contains 7,358 individual scans at various poses of 396 subjects.
The dataset has a number of full profile scans which are correctly aligned and contribute directly to the associated mesh geometry.
The dataset in its raw form contains a significant number of mislabeled scans, which are identified and corrected based on alignment error using the proposed algorithm.
The average point to surface distance between the aligned scans and the produced geometries is one half millimeter.
\end{abstract}

\section{Introduction}\label{INTRODUCTION}
We propose a 3D alignment and template warping pipeline suitable for merging multiple "single shot" 3D scans of human faces from various angles and estimating parameters of a 3D Morphable Model (3DMM)~\cite{mcare3} for the given subject.
The proposed pipeline is capable of handling large datasets that include significant numbers of subject labeling errors.
We consider 3DMM coefficent estimation a specific form of template warping, where an ideal mapping is found between a template surface (the "mean face") and a target (the subject's face geometry).

The tasks of alignment and template warping are usually approached separately with very different techniques, however we solve the alignment and surface registration problems simultaneously by alternating between a rigid alignment of the scan data to the template mesh and a warping of the template to the scan data. By combining the two processes of alignment and surface registration, we simplify both problems while achieving robust and pose invariant results on unstructured data. 
	
		
We show quantitative and qualitative evidence for the accuracy in alignment between scans as well as the accuracy of the resulting parameterized 3D facial meshes described by the estimated 3DMM parameters.

\begin{figure}[t]\label{_table}
\begin{tabular}{cc}
\includegraphics[scale=0.09]{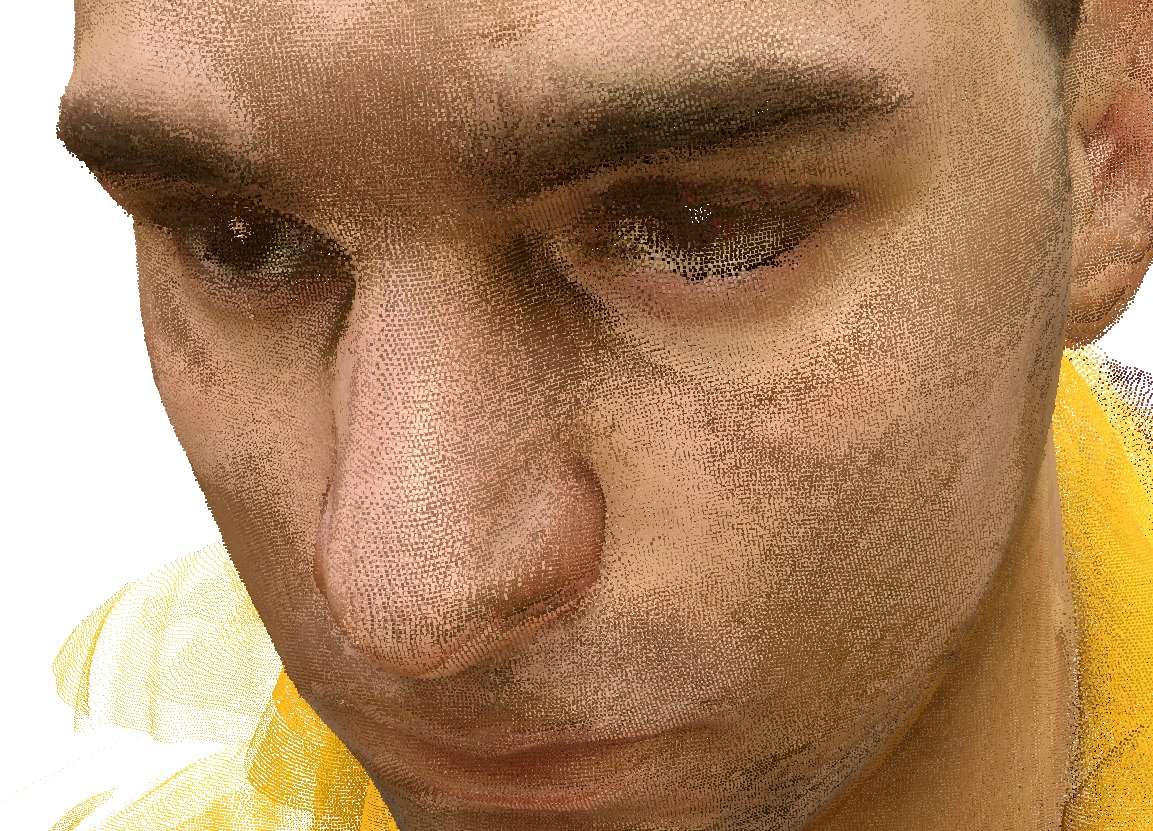} & \includegraphics[scale=0.09]{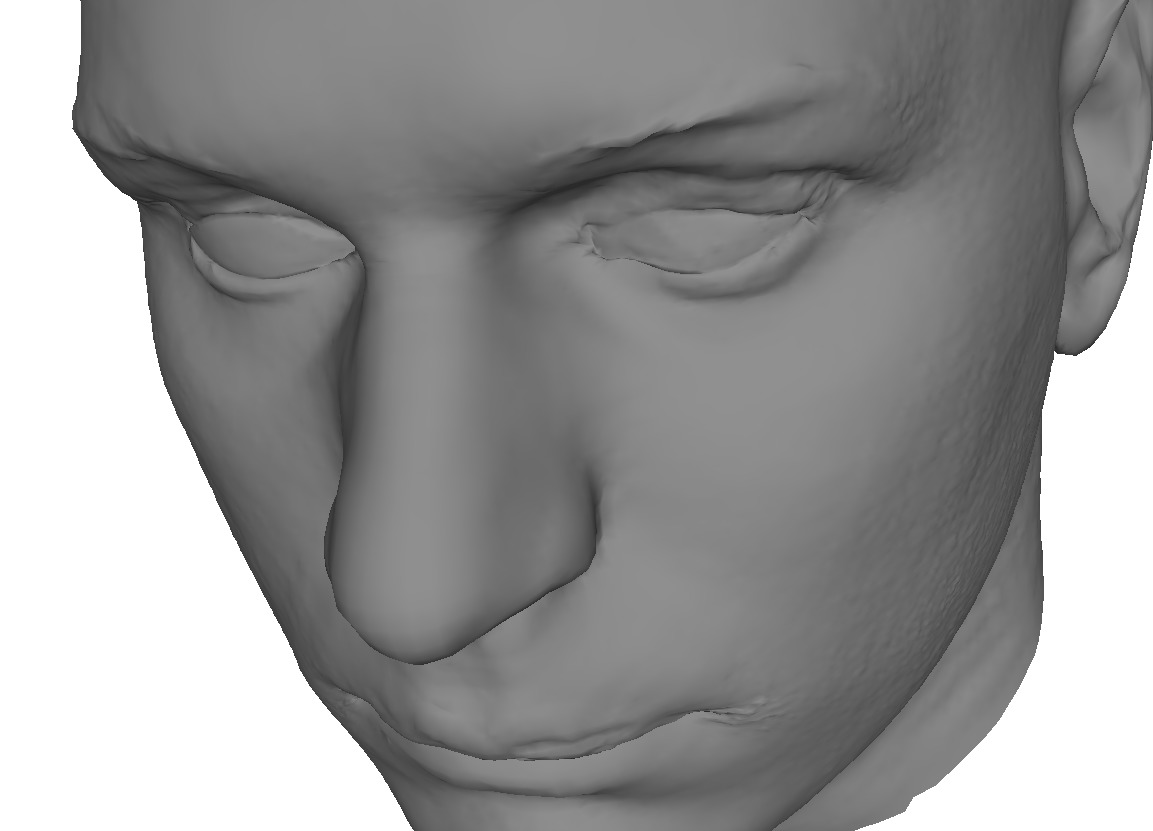} \\
\includegraphics[scale=0.09]{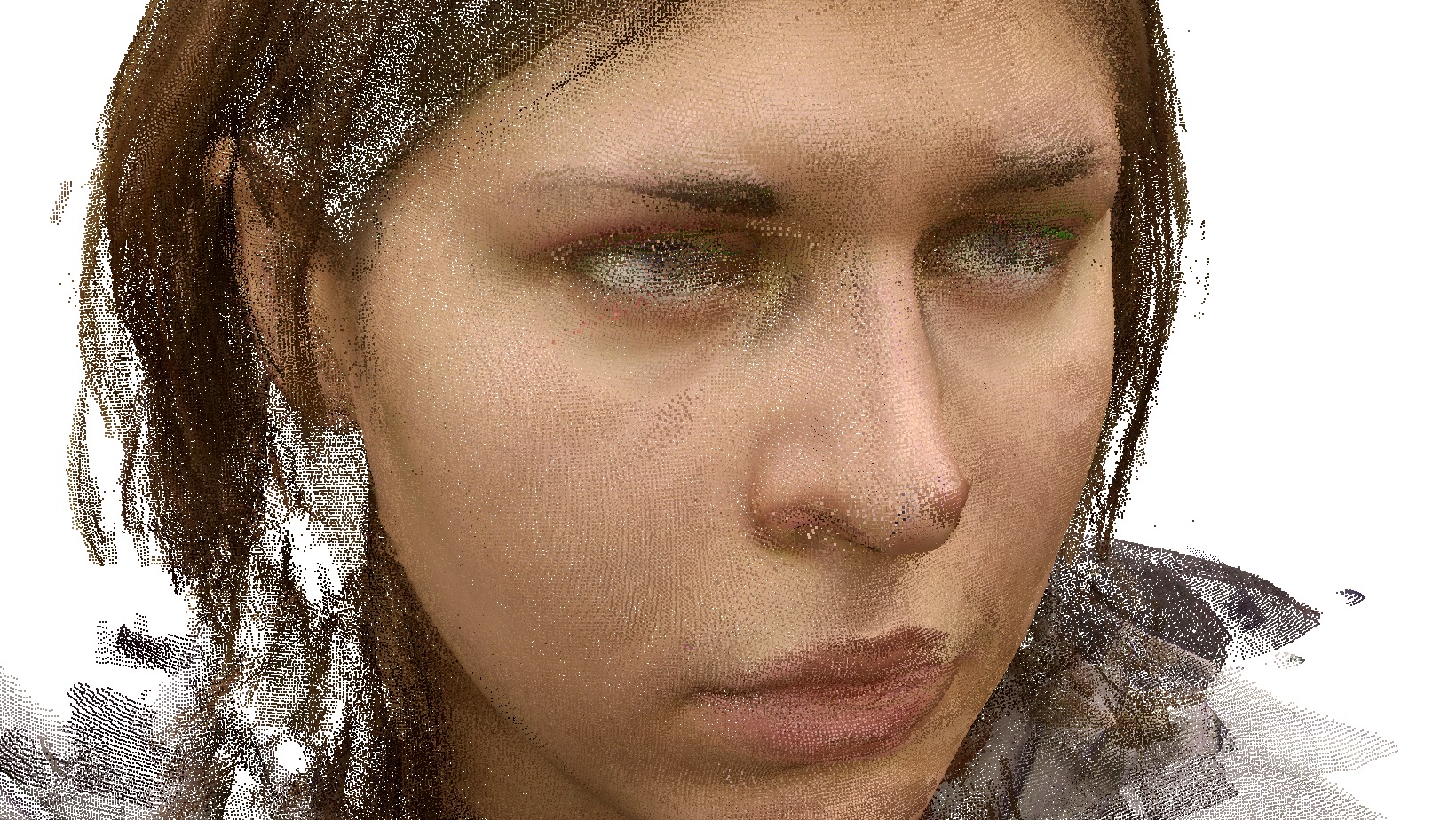} & \includegraphics[scale=0.09]{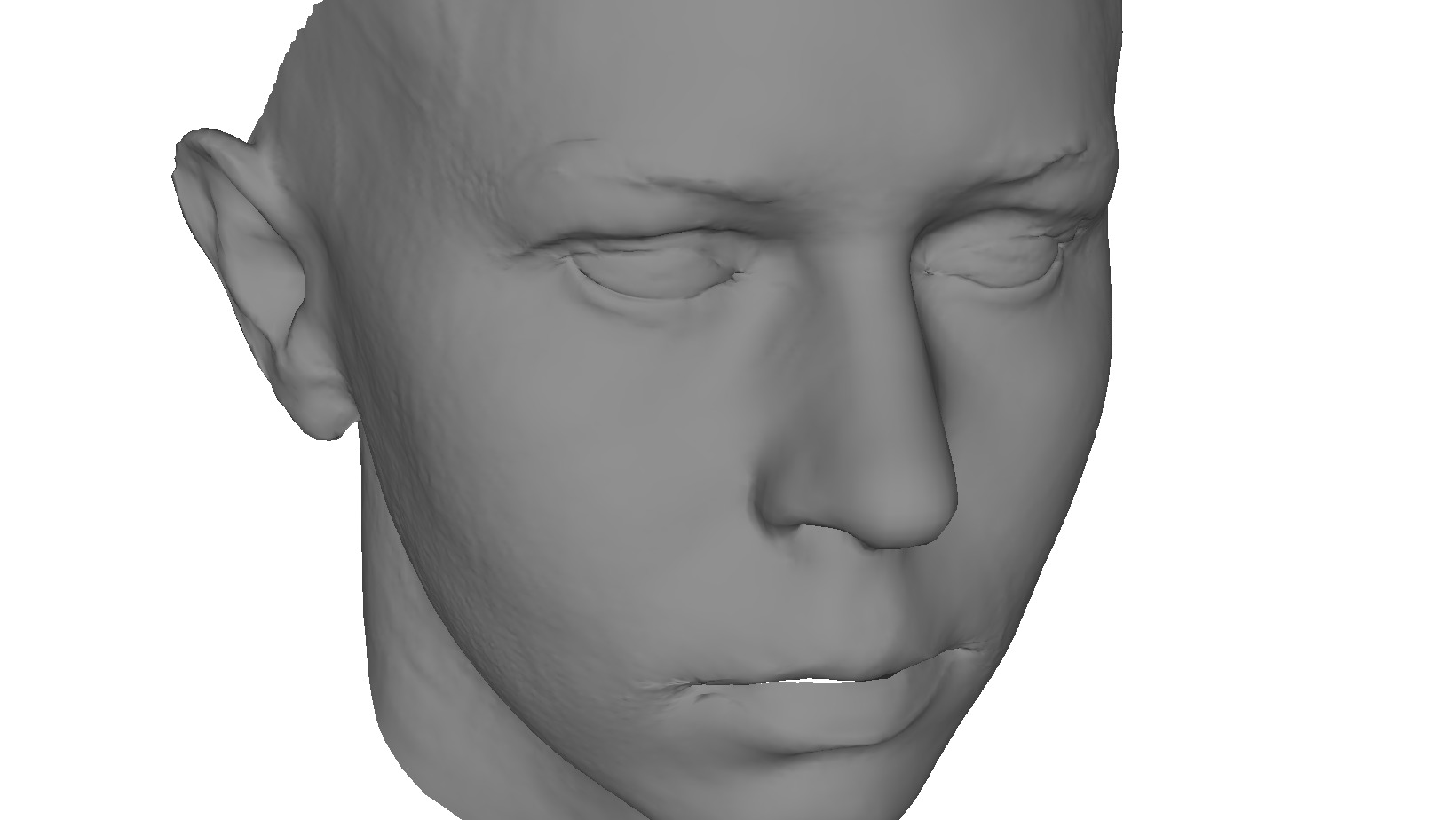} \\
\includegraphics[scale=0.09]{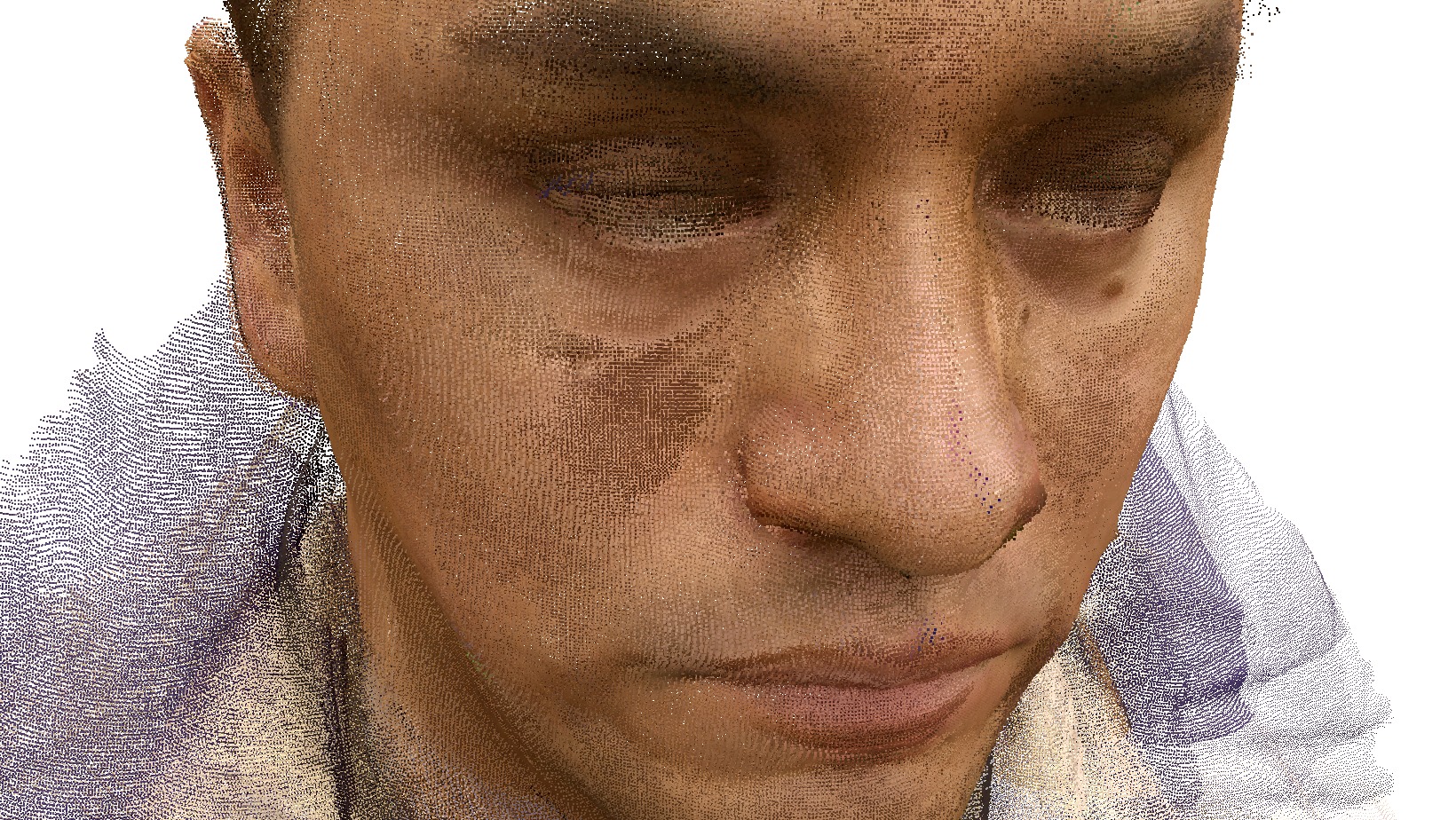} & \includegraphics[scale=0.09]{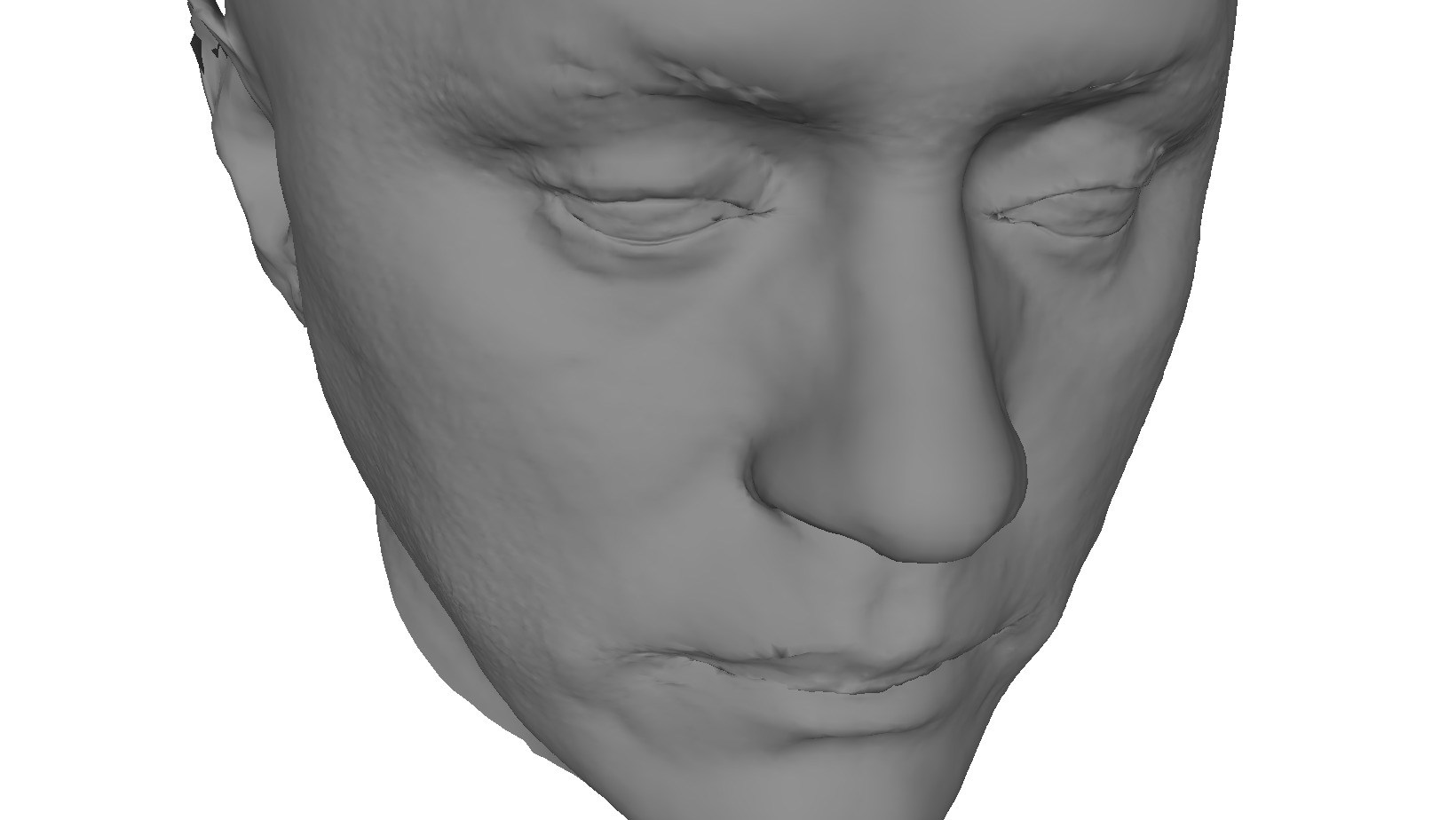} \\
\includegraphics[scale=0.09]{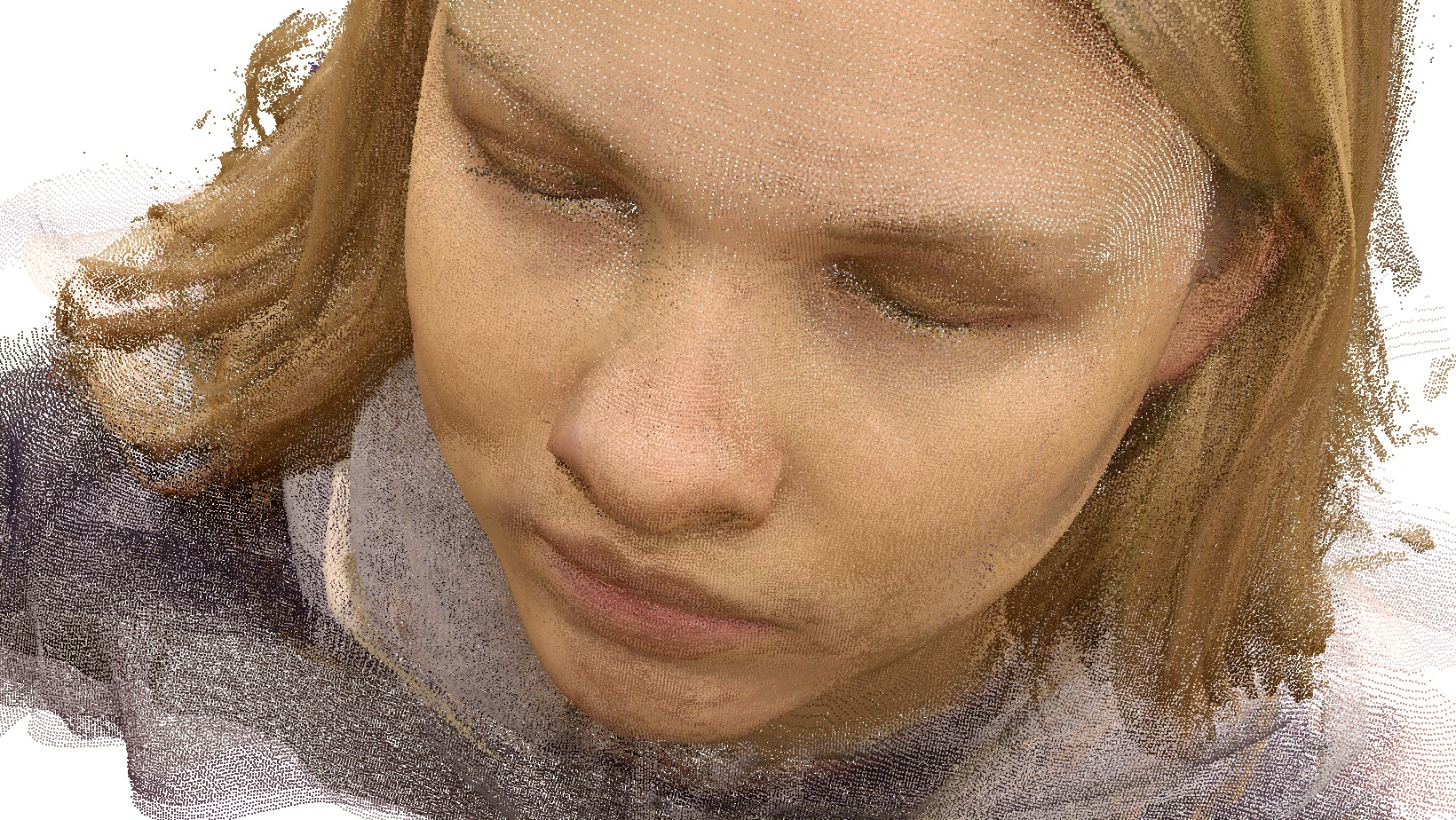} & \includegraphics[scale=0.09]{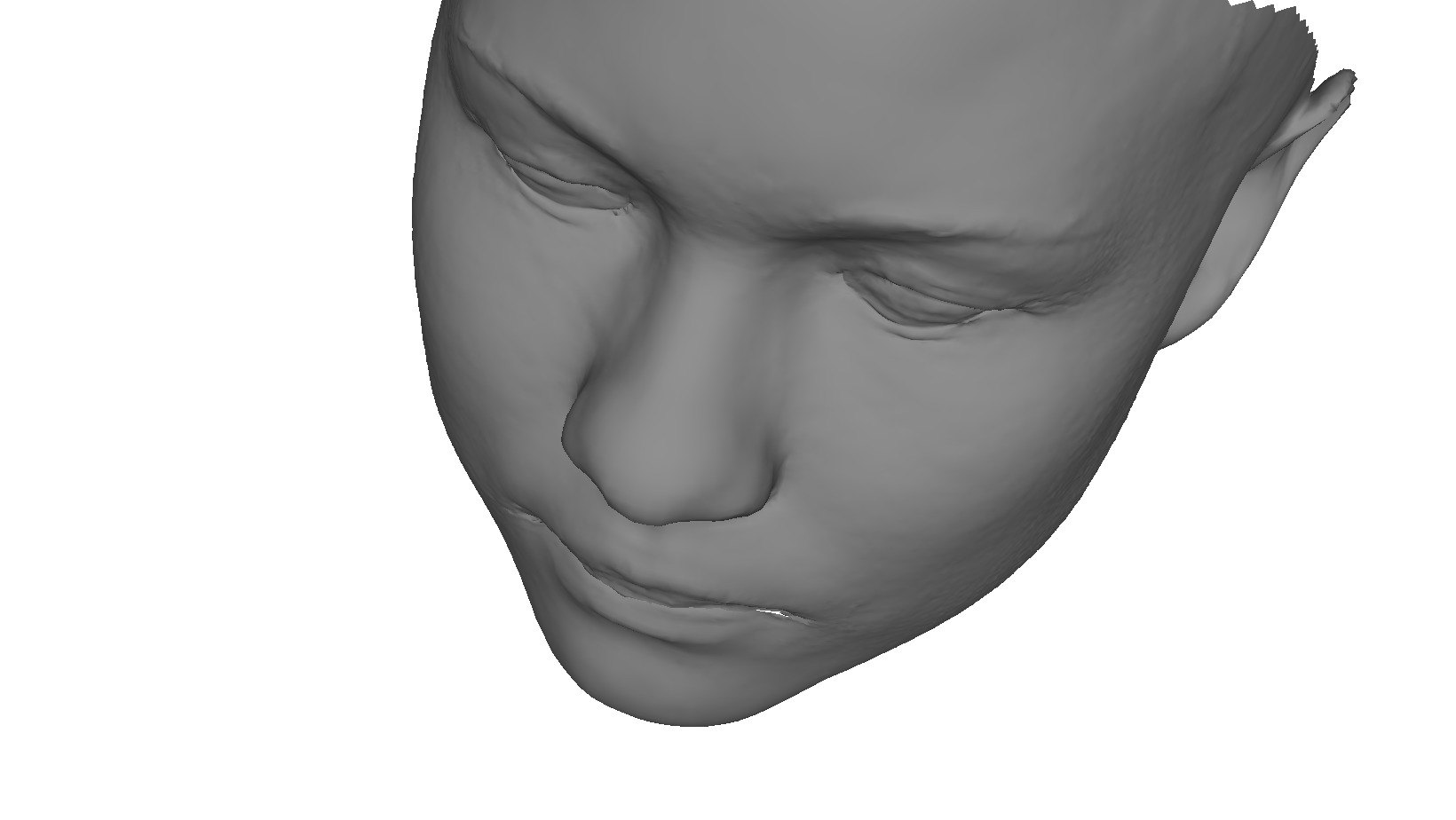} \\
\end{tabular}
\centering
\caption{Aligned scans [left] and mesh geometry [right].}
\end{figure}

\section{Contributions}\label{Contributions}
\begin{itemize}
\item A fully automatic, end-to-end pipeline for alignment of 3D face scans and per-subject 3DMM coefficient estimation.
\item Investigation into the effect of the proposed iterative refinement method on point cloud alignment and mesh geometry estimation.
\item Aligned 3D scans available through The University of Notre Dame, for the benefit of the community.
\end{itemize}

\section{Related Work}\label{Related_Work}

With the rise of commercial scanners, 3D information is more widely available than ever before. In order to fully capture a 3D object using general-purpose commercial scanners, the merging of multiple scans is often required. We address the issue of aligning multiple 3D facial scans from arbitrary unknown pose by generating a mesh representation of the subject which is refined in tandem with alignment. Our method has the added benefit of producing a warped template mesh which is closely aligned with the scan data.

Because the pipeline produces both aligned scan data and a warped template mesh, we briefly go over both the problem of alignment and the problem of template warping. The goal of template warping is to preserve the semantic meaning of each vertex in the template while expressing the geometry of the target data. As far as we know, there is no previous work which jointly addresses both scan alignment and template warping. However each problem separately has been studied for many years and has a rich history.

Rather than using unconstrained template warping, we employ a 3D Morphable Face Model, which uses a set of 199 PCA components to describe observed variation in human face geometry. 3DMM-based approaches have enjoyed much success over the past decade in the contexts of various computer vision and computer graphics applications~\cite{mcare1,mcare2,mcare3}. Typical approaches to producing a 3DMM rely on on surface registration techniques that assume a single template and target mesh. Allen et al.~\cite{Allen} apply an affine transform to the template vertices that minimize an error consisting of three parts: the distance from each mesh vertex to the target mesh, the similarity of the transform between connected vertices, and the error between ``marker points'' that have known positions on the template and target. The error quantity described is used to register a full body template with a full body scan. Amberg et al.~\cite{Amberg} use an error term consisting of the same three parts as Allen~\cite{Allen}, and solve for the error using non-rigid iterated closest point (ICP). Both techniques preserve local structure by gradually decreasing the significance of the stiffness term, where stiffness here means the portion of the error term that penalizes differences in affine transforms between connected vertices. Amberg's approach is employed to great success by Booth et al.~\cite{Booth} in producing a 3DMM learned from 10,000 high resolution 180 degree facial scans. The work of Booth showed how more data directly resulted in a more accurate and expressive 3DMM which outperformed the state of the art.

Facial scan alignment is typically reduced to a problem of landmark localization. There exists a rich diversity of landmark estimation algorithms which use 2D and 3D information. Kazemi et al.~\cite{Kazemi} use an ensemble of regression trees to achieve super real-time localization on the order of one millisecond per image.  Merging the problem of face detection, landmark localization and pose estimation, Xiangxin et al.~\cite{Zhu} use tree structured models and outperformed the state of the art on serveral benchmarks. When depth information is available, Fanelli et al. \cite{Fanelli} show that accuracy can be improved, while maintaining speed. Xiangyu et al.~\cite{Xiangyu} employ a fundamentally different approach by building a cascade convolution neural network (CNN) to directly estimate camera parameters and the 3DMM coefficients using 2D images as input.

The proposed alignment method begins by using sparse localized landmarks, but as dense 3D information is available and real time performance is not necessary, additional steps are taken to refine the initial alignment by registering each scan to the subject-specific mesh geometry. Mesh geometry is computed by finding a set of 3D offsets that express the local difference between the scans and the base mesh. These offsets are used at first to warp a 3DMM using precomputed PCA components. Direct (unconstrained) mesh warping without the PCA model is performed as a final step by a method similar to the one described by Arberg~\cite{Amberg}. The major geometry variations are described by the 3DMM warping, while the direct approach is able to account for smaller, finer details not represented by the 3DMM\@. The significant warping that occurs using the 3DMM PCA components removes the need for a decreasing stiffness parameter when estimating the direct warping.

\section{Data}\label{Data}
The Notre Dame facial scan dataset is comprised of 7,358 3D laser scans from 396 individuals, averaging 18.58 laser scans per individual. Each scan is composed of an 8-bit RGB image with resolution 640x480, and a set of $(x,y,z)$ coordinates corresponding to each pixel of the image. Each laser scan captures only a portion of the subject's face at different yaw and pitch. The $(x,y,z)$ coordinates are in a camera-centric coordinate system, i.e.\ they are not aligned with respect to the subject's face. Laser scans are grouped by subject, however a significant portion of scans are incorrectly labeled.

\section{Preprocessing}\label{Preprocessing}
Two processing steps are done which remove unwanted surfaces from the scans and reduce the size of the laser scans. The first is a crop around the face, as we are not concerned with scanned surfaces which are not part of the face. The second step is to detect and remove hair, which is highly variable between scans making it hard to capture in a template and problematic for alignment. Once a subject's scans are fully aligned, the points removed by preprocessing are restored to maintain completeness.

\subsection{Face Detection and Cropping}
Faces are detected using a Convolutional Neural Network (CNN)-based face detector implemented as part of the dlib~\cite{dlib} software library. The facial bounding box is expanded by 30 pixels to include ares of the face which often lie outside of the detection box, like the upper forehead and lower chin. When multiple faces are detected the larger bounding box is used, as the subjects' faces typically occupy most of the frame. In the rare cases where no face is detected the scan is discarded.

\subsection{Removing Hair}

\subsubsection{Computing Normals}
The first step in the hair detection process is to compute the surface normals for each vertex in the point cloud. Computing the surface normal ($\mathbf{n_v}$) for a vertex ($\mathbf{v}$) is done in three steps. Initially, a neighborhood of the 30 vertices which have the minimum Euclidean distance to $\mathbf{v}$ are obtained. Finding the neighborhood for a vertex is done in $\mathcal{O}(\log{n})$ time by storing all vertices in a KD-tree. Next, the vertex and its neighborhood are translated so that their centroid is at the origin. When the translated point of the vertex $\mathbf{v}$ and its neighborhood are represented as columns of a matrix ($A$) the normal of the plane which minimizes the squared euclidean distance is the left singular vector of $A$ corresponding to the singular value of $A$ which is of smallest magnitude.
\subsubsection{Normal-Based Filtering}
\begin{figure}
 \centering
 \includegraphics[scale=.2]{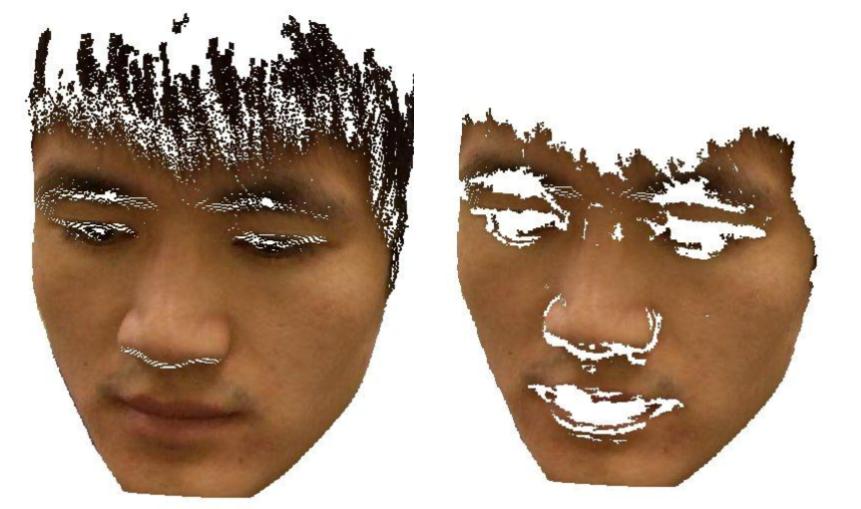}
 \caption{Laser scan before [left] and after [right] the hair filtering process. Removal of points around the lips, eyes, and base of the nose is typical due to the increased curvature in these areas.}
	\label{fig:normfilt}
\end{figure}

Hair is removed by a normal-based filtering approach, which removes points with surface normals that are significantly different in angle from the surface normals of points in the immediate neighborhood. Each point's neighborhood consists of the 10 closest points, ordered by Euclidean distance. If the mean angular difference between a point's surface normal and the surface normals of its neighborhood is greater than a fixed angular threshold, the point is determined to be part of a rough surface and it is removed. A threshold of eight degrees was found to work well in practice. In addition to hair, facial features with high curvature such as the lips, eyes, and the base of the nose are sometimes removed from scans as shown in Figure~\ref{fig:normfilt}.  While this is not a desired outcome, we did not observe it affecting the alignment accuracy in practice.

\subsubsection{Density Based Clustering}
After the normal-based filtering step, small, disconnected patches of hair may still be present in areas where the hair is locally flat. To remove these disconnected regions, density-based spatial clustering (DBSCAN)~\cite{dbscan} is performed on the scan. Experiments show that a maximum distance of 1.5 mm between neighbors of a cluster is sufficient to group all facial points together while excluding other regions. Once clustered, The face if easily identifiable as the largest cluster and the remaining clusters are discarded.

\section{Initial Pose Estimation}\label{Initial_Estimation}
An initial estimate of subject pose is generated using the face reconstruction method of Crispell et al.~\cite{Crispell} based on sparse 2D landmarks estimated using the image. Using estimated camera and geometry, a dense set of correspondences is found between the mean face and the subject's. From the dense correspondences the optimal rigid transformation is computed.

When the aforementioned localization fails, each landmark is matched with its known 3D location on the mean face, and for every combination of three correspondence pairs, a rigid transform is computed. The transform which minimizes the point-to-mesh distance is selected as the initial transform. This method for alignment is extremely robust as it only requires three correct landmarks. However, since this method is less accurate at estimating pose, the scans initialized in this way are not used to compute geometry offsets for the first iteration of mesh generation. After the first iteration, the alignment converges with sufficient accuracy to allow the scans to function identically to those initialized in the standard way.

\section{Geometry Estimation}\label{Geometry}
For each laser scan a set of mesh vertex positions are calculated which approximately minimize point to surface distance between the scan and the mesh. The vertex coordinates are encoded as offsets $O$ from the mean face mesh $\bar V$. The vertex offsets are used to estimate a new set of PCA coefficients, optionally with a detail vector $\delta$. The detail vector makes up the difference between the optimal vertex positions and those generated by the PCA coefficients.

\subsection{Computing Scan Level Offsets}\label{section:offsets}
In this section the process for computing the mesh offsets for a single scan is defined. The total set of offsets will be used for the rest of this section to define how the geometry is updated. A single offset matrix $O_i$ is of the same shape as $\bar V$. $O_i$ is generated from the current mesh $V_i$ and a single aligned laser scan $P_i$ such that the following equation is minimized:

\begin{equation}\label{eq:epsilon}
d(O_i + \bar V, P_i)
\end{equation}

where $d(g,p)$ computes the point to surface distance between the mesh defined by geometry $g$ and point cloud $p$.

Each row of the offset matrix $O_i$ has a corresponding vertex in the mean mesh $\bar V$ and is computed independently from other vertices. Because of this independence we can define the process for estimating a single vertex $v$ once and then apply this method to all vertices.

\begin{figure}[h]
\centering
\includegraphics[scale=.55]{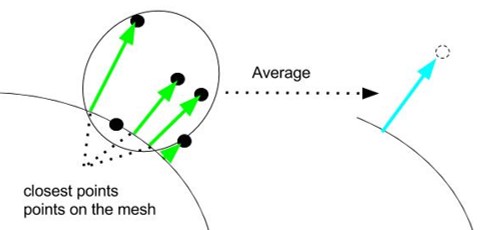}
\caption{Computing the relative offset.}
\label{fig:offsets}
\end{figure}

First, each point in $P_i$ is paired with its closest point on the current mesh surface. Of the paired points on the mesh's surface, the three which are closest to the vertex $v$ are used to compute the offset (Figure~\ref{fig:offsets}). Once the three points of $P_i$ are found, the relative offset of $v$ is set as the mean distance from the paired point on the mesh to the points of $P_i$. As a final step, the vertex offset is calculated as the relative offset plus the vertex offset of $v$ in the current mesh. 

As the scan level offsets are meant to be a low level representation, it is advantageous to not compute offsets for holes or empty regions. To prevent offsets from being calculated for erroneous regions, a distance check is performed; if the distance between the vertex and the three closest point on the mesh is greater than a threshold, the vertex is not assigned any offsets. Regions without any offsets, such as holes, are handled robustly through the use of the PCA model, which accounts for the full facial geometry.

\subsection{Solving for the Structure Coefficients}
The scan level offsets are used to warp the mesh by computing the coefficients $A$ of the structure vector $\alpha$. The vector $\alpha$ is a dimensionality reduction of the space of $V$, found using PCA and trained using a series of synthetic faces. The methods for computing $\alpha$  similar to those used by Crispell et al.~\cite{Crispell}, but no projection to 2D is needed in this case. The benefit of using the structure vector is the robust estimation of regions which are not observed~\cite{Crispell,Booth}.

Two methods are used for computing $A$ to satisfy the constraints of $O_1 O_2 \hdots O_I$. The first method, shown in equation \ref{eq:svd1}, gives equal weight to each set of offsets. The drawbacks of this method are the memory and computation resources required to solve such a large set of equations. As the number of equations grows linearly with respect to both the size of the mesh and the number of offsets, it was found that in practice there were sometimes as many as two million linearly independent equations which took several minutes to solve using state of the art commercial off the shelf hardware.

\begin{equation}\label{eq:svd1}
\begin{bmatrix}
  A \\
	A \\
  \vdots \\
  A \\
\end{bmatrix}
\cdot
\begin{bmatrix}
   \alpha \\
\end{bmatrix}
=
\begin{bmatrix}
   O_1 \\
   O_2 \\
   \vdots \\
   O_I \\
\end{bmatrix}
\end{equation}

The second method (Equation~\ref{eq:svd2}) solves for $A$ using the mean of the set of offsets $\bar O$. To account for the increased accuracy which comes from averaging many offsets, a weight vector $w$ is added. $w$ has values equal to the number of point clouds used for computing the mean offset at each mesh vertex. The weight vector $w$ is non uniform as offset matrices typically have some vertices which have no value. Similarly there are some vertices which may not be observed at all. The position of these vertices does not contribute to the solution as the weight at these locations is zero. Vector $w$ is applied to $A$ and $\bar O$ using the element-wise multiplication, shown in equation~\ref{eq:svd2} as $\odot$. This second method is much faster than the first as it depends only upon the size of the mesh while producing meshes of comparable quality.

\begin{equation}\label{eq:svd2}
w \odot \alpha A = w \odot \bar O
\end{equation}

Because in practice accuracy was comparable using both techniques the second method was selected as it has more desirable computational requirements.

\subsection{Solving for Detail}
Once close alignment is achieved using the PCA coefficients, it is advantageous to introduce a detail vector $\delta$. Despite the expressiveness of the PCA model, there are some facial geometries which cannot be fully realized using only the estimated coefficients. These are typically fine details such as wrinkles that the linear PCA-based model is incapable of reproducing.

\begin{equation}\label{st}
\exists V, \forall A  \mid V \neq \bar V +  \alpha A
\end{equation}

The detail vector $\delta$ closes the gap between the mesh parameterized by $A$ and the geometry $V$.

\begin{equation}\label{eq:abd}
V = \bar V + \alpha A  + \delta
\end{equation}

The detail vector is calculated as the difference between the mean offsets and the geometry produced by the PCA model and coefficients.

\begin{equation}\label{eq:d}
\delta = \bar O - \alpha A
\end{equation}

The detail vector is only estimated after the PCA coefficients have already been computed. Additionally, the detail vector is not used until the estimated geometry has already been refined several times. The detail vector is computed for all vertices which have at least one offset $V = \bar V + \bar O$. When there is is no offset information available for a vertex, positions are estimated using only $\alpha$, which produces reasonable estimates.

\begin{figure}[h]\label{detail_table}
\begin{tabular}{c | c}
\includegraphics[scale=.33]{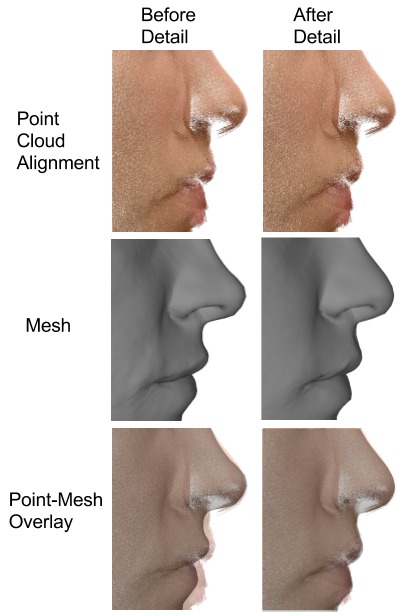} &
\includegraphics[scale=.33]{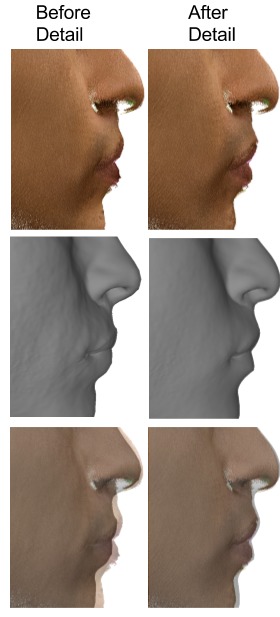}
\end{tabular}
\centering
\caption{Meshes and point clouds before and after applying the detail vector.}
\end{figure}

\section{Point Cloud Alignment}\label{ICP}
Point clouds are aligned to each other indirectly by independently aligning each to a mesh which best expresses the shape of the subject's face at any given iteration. To align a point cloud to a mesh a modified Iterated Closest Point (ICP)~\cite{icp} algorithm is applied. 

The ICP alignment process operates as follows. Initially the points of the scan are matched to their closest points on the mesh. From the corresponding mesh points an optimal rigid transform is computed and applied to the scan point cloud. The closest points on the mesh are reestablished and the process is repeated iteratively until convergence or a maximum number of iterations is reached.

In practice, two modifications are made to this nominal fitting approach. First, before alignment each point cloud is randomly downsampled to one tenth the original size. If was found that down sampling has a negligible effect on accuracy while greatly improving performance. Second, a subset of the face is used for alignment. The crop was designed for two reasons: to remove the ears, which due to their thin and variable structure are often poorly aligned with the mesh, and to remove the neck due to its variable pose relative to the head. In the typical case, ICP converges well within 50 iterations, however 100 iterations are allowed for the pathological case.

\section{Iterative Processing}\label{Iterative}
After the preprocessing (\ref{Preprocessing}) and initial pose estimation (\ref{Initial_Estimation}), the iterative refinement phase begins. The iterative refinement is divided into three phases; each phase receives a set of aligned point clouds, updates the alignment and generates a new mesh. Phase I begins with the worst alignment and performs geometry estimation (\ref{Geometry}) and alignment (\ref{ICP}) four times. During phase I the geometry is the least accurate, which is why the detail parameter is not used. Experiments show that after four iterations the mean distance of the point clouds to the mesh plateaus. 

The detail coefficient is introduced to further reduce error in Phase II\@. The detail coefficient warps the mesh to fit the intricate details of the subject's face. During phase II the geometry and alignment are iteratively estimated three times. After three iterations error plateaus again.

Once phase II completes meshes are typically accurate enough to correctly identify mislabeled scans. The mislabeled scans are removed and relabeled using the techniques described in Section~\ref{Relabel}. After relabeling, the mesh geometry is estimated a final time, And using this final geometry the final point cloud alignment is found.

\begin{figure}[t]
\centering
\includegraphics[scale=.5]{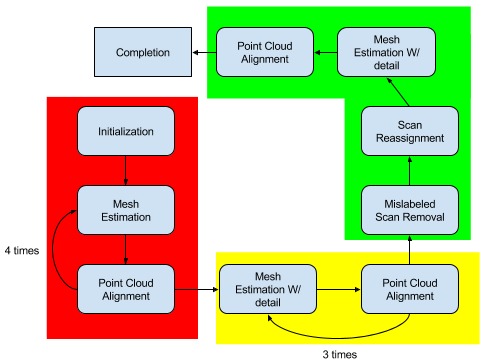}
\caption{Phase I [red], phase II [yellow], and phase III [green] of the alignment pipeline.}
\end{figure}

\section{Relabeling}\label{Relabel}
Mislabeled scans are identified using only the 3D positions of the point cloud and shape of the associated mesh. Identifying mislabeled scans is done by thresholding the mean squared point to surface distance between the scans and the meshes. A threshold of 1.1mm for the mean squared distance to the mesh was found to be sufficient in removing most mislabeled meshes while having a minimal effect on correctly labeled faces.

Once the mislabeled point clouds are removed, a more standard approach is taken to relabeling. Each image is passed through a Convolutional Neural Network (CNN) trained to generate identity encodings on images of human faces~\cite{Parkhi15}, which produces a unit vector of dimension 4096. A subject's encoding is taken to be the mean of the encoding of all the subject's images. Classification is done by solving for each outlier image's match using K-nearest neighbors.

As we have access to the 3D geometry of the subject as well as the 3D scan corresponding to each image, we enforce additional constraints on the matches. Once a relabeling is hypothesized, the scan is aligned to the new subject's mesh using the ICP method described in section~\ref{ICP}. If the mean squared error is below the threshold of 1.1mm the relabeling is accepted.

\section{Error Analysis}\label{Error}
\begin{figure}
\includegraphics[scale=.55]{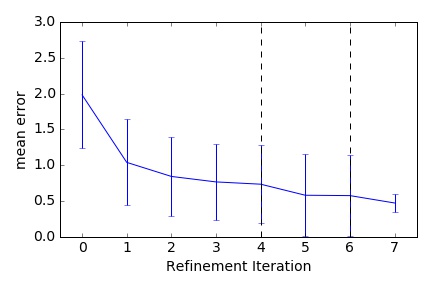}
\caption{The mean error and variance as measured by the distance from each scan point to the closest point on the mesh. The vertical lines at iteration four and six mark the introduction of the detail parameter and removal of mislabeled data respectively.}
\label{fig:lin_scale}
\centering
\end{figure}

To measure error we computed the mean distance from each scan to the closest point on the subject's mesh. We found that error decreases at every iteration. Graph~\ref{fig:lin_scale} shows the mean error and the variance in error across all subjects. Notice that the variance drops very steeply once mislabeled images are removed or relabeled. 

\begin{figure*}
\begin{tabular}{>{\centering\arraybackslash}p{1.5cm} | m{8mm} m{18mm}m{18mm}m{18mm}m{18mm}m{18mm}m{18mm} m{18mm}m{18mm}m{18mm}}

Scan&
&
\includegraphics[width=23mm]{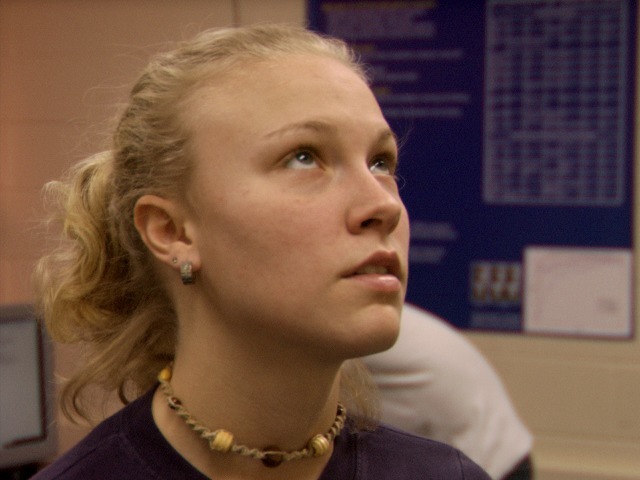} &
\includegraphics[width=23mm]{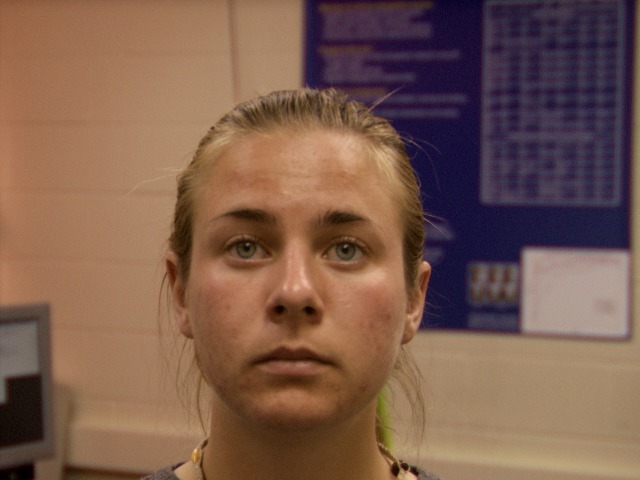} &
\includegraphics[width=23mm]{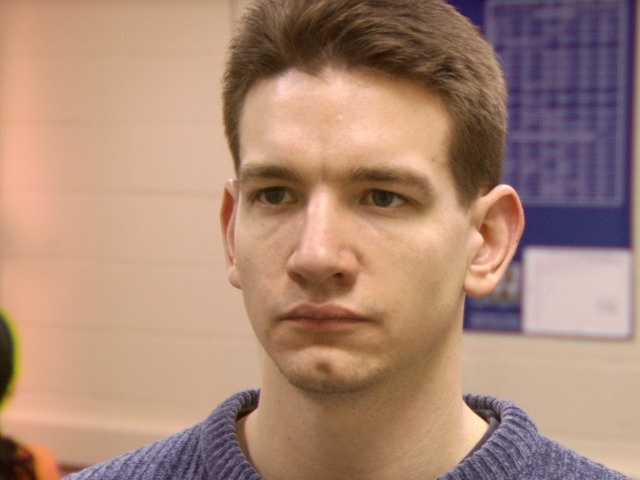} &
\includegraphics[width=23mm]{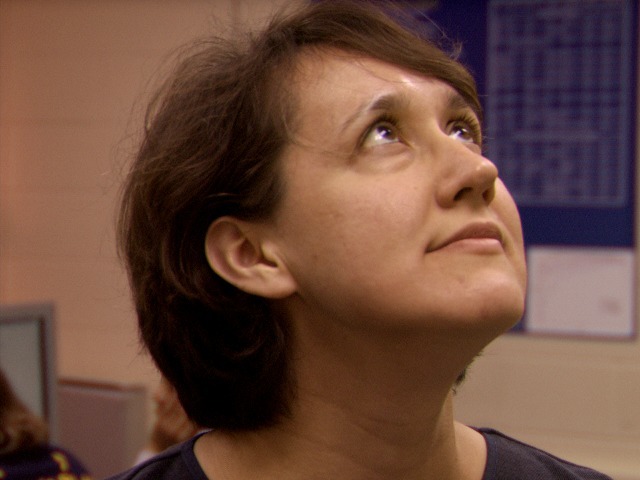} &
\includegraphics[width=23mm]{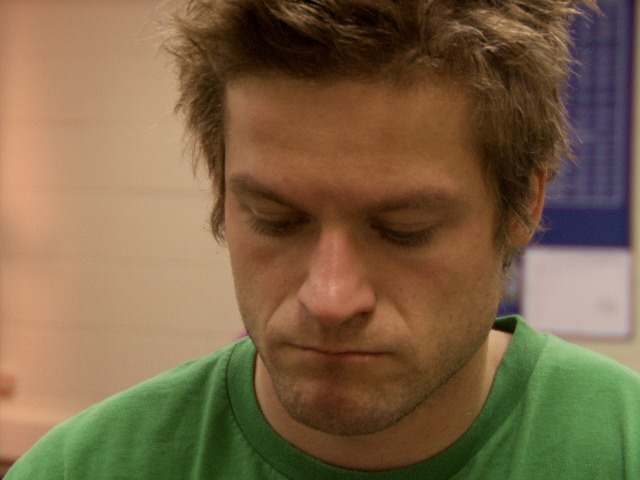} &
\includegraphics[width=19mm]{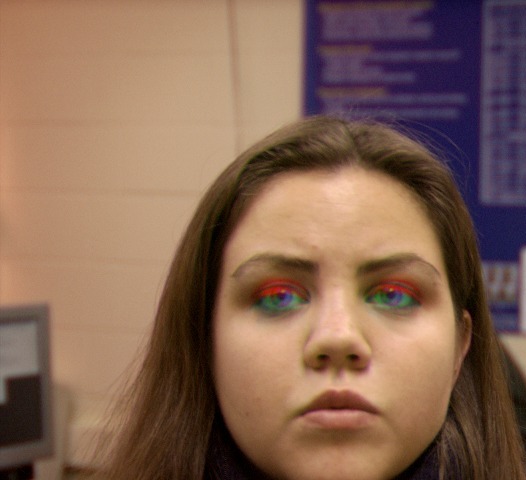} \\

\noalign{\vskip 2mm}  
\hline
\noalign{\vskip 2mm}  

Initial Error Full &
\multirow{2}{*}{\includegraphics[scale=0.35]{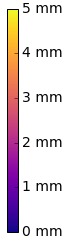}} &
\includegraphics[width=23mm]{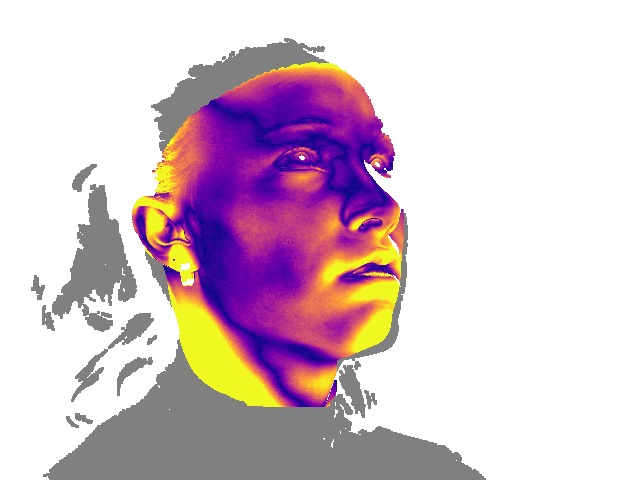} &
\includegraphics[width=23mm]{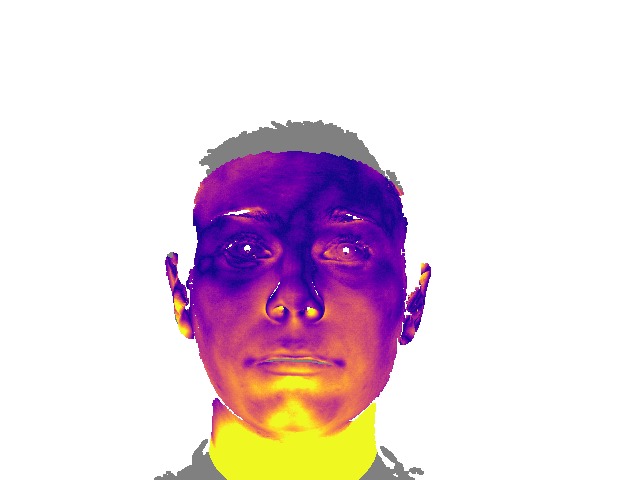} &
\includegraphics[width=23mm]{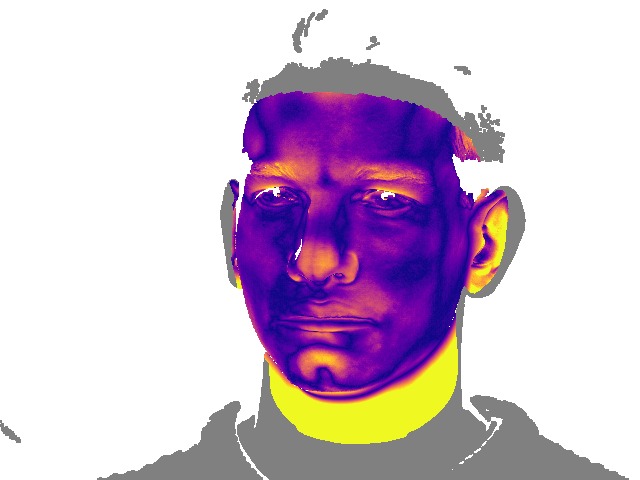} &
\includegraphics[width=23mm]{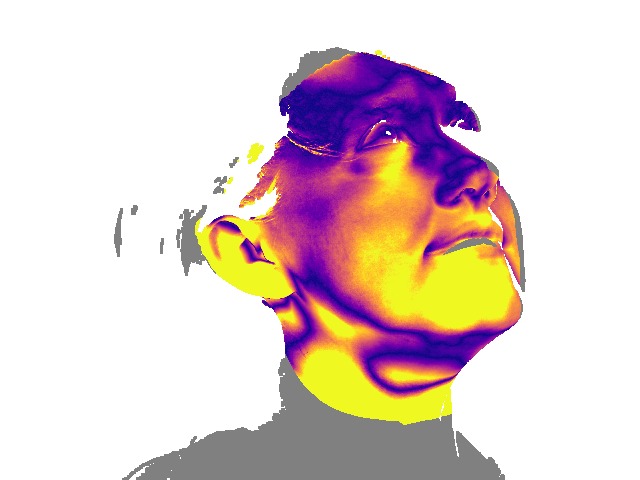} &
\includegraphics[width=23mm]{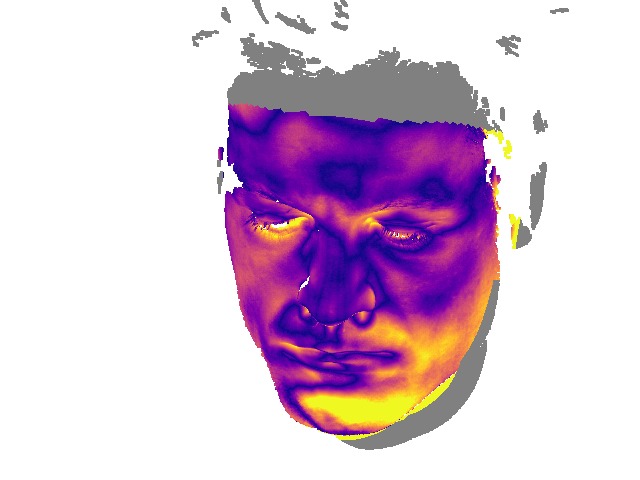} &
\includegraphics[width=23mm]{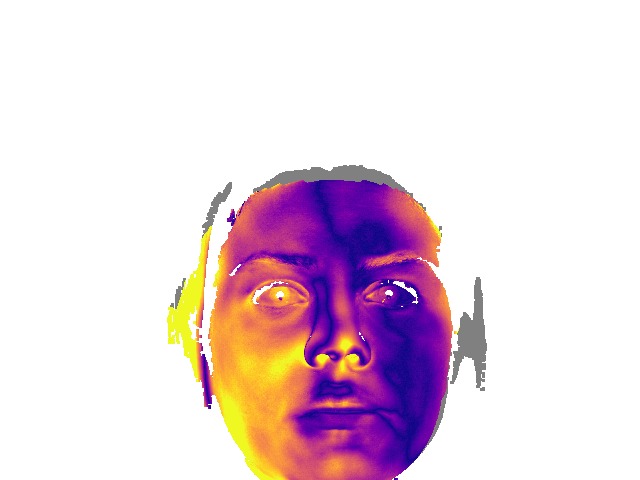} \\

Initial Error &
&
\includegraphics[width=18mm]{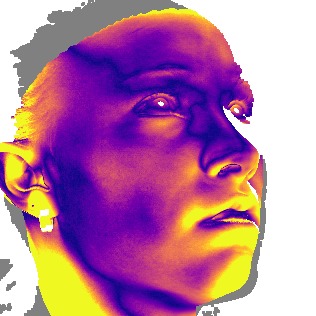} &
\includegraphics[width=18mm]{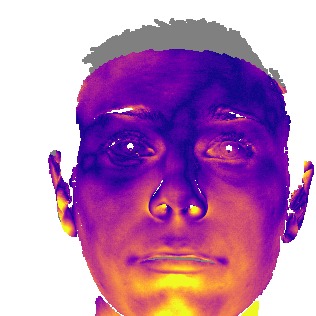} &
\includegraphics[width=18mm]{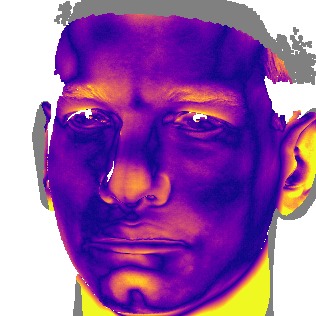} &
\includegraphics[width=18mm]{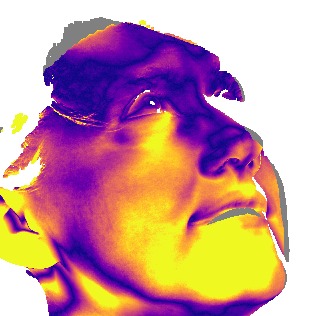} &
\includegraphics[width=18mm]{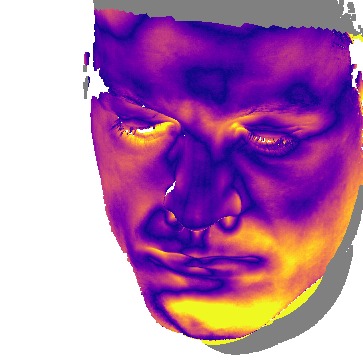} &
\includegraphics[width=18mm]{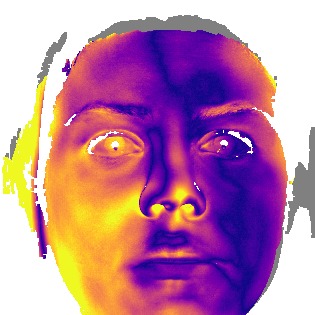} \\

Initial Mesh&
&
\includegraphics[width=18mm]{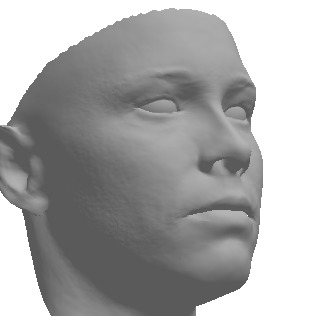} &
\includegraphics[width=18mm]{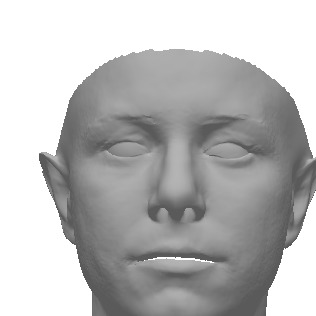} &
\includegraphics[width=18mm]{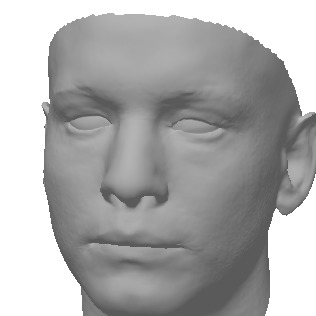} &
\includegraphics[width=18mm]{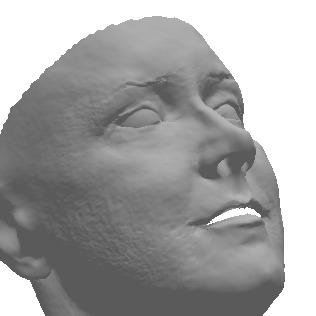} &
\includegraphics[width=18mm]{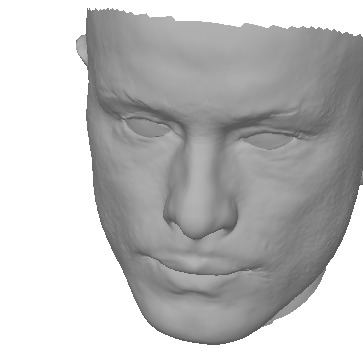} &
\includegraphics[width=18mm]{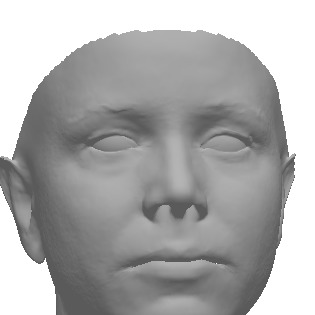} \\

Initial Mesh Close Up&
&
\includegraphics[scale=0.36]{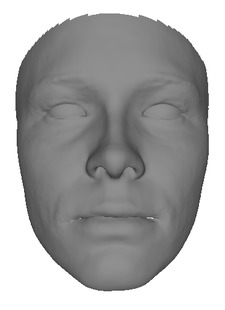} &
\includegraphics[scale=0.36]{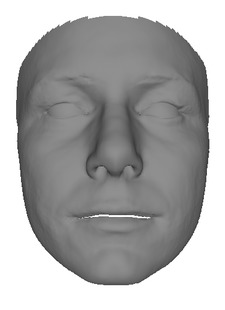} &
\includegraphics[scale=0.36]{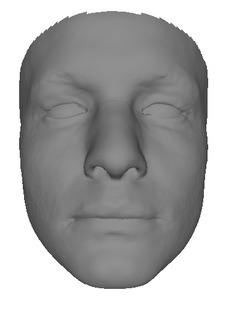} &
\includegraphics[scale=0.36]{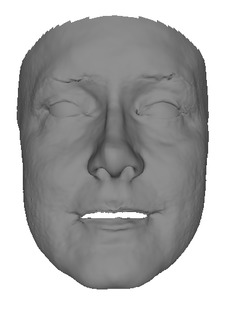} &
\includegraphics[scale=0.36]{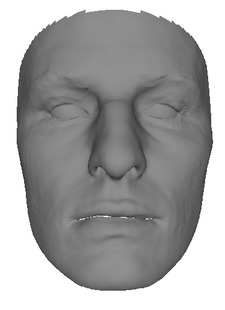} &
\includegraphics[scale=0.36]{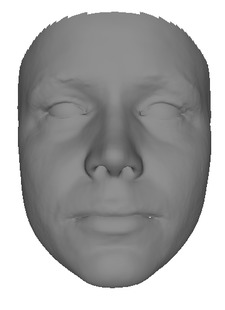} \\

\noalign{\vskip 2mm}  
\hline
\noalign{\vskip 2mm} 

Final Error Full &
\multirow{2}{*}{\includegraphics[scale=0.35]{gi2/cbar_crop.jpg}} &
\includegraphics[width=23mm]{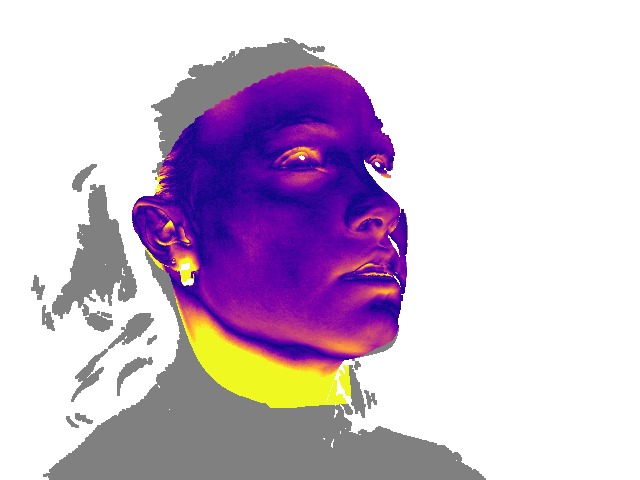} &
\includegraphics[width=23mm]{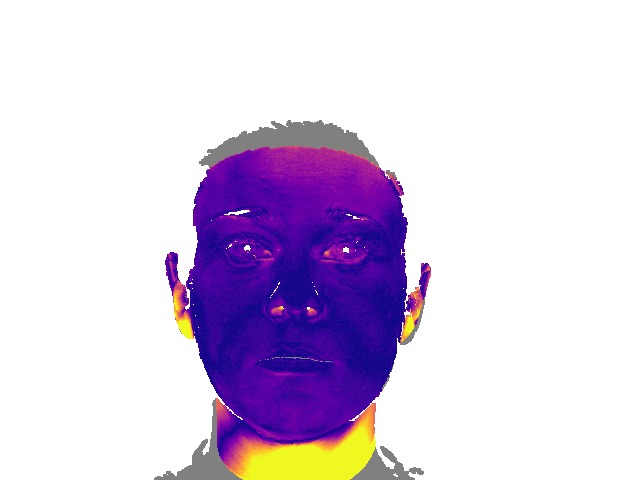} &
\includegraphics[width=23mm]{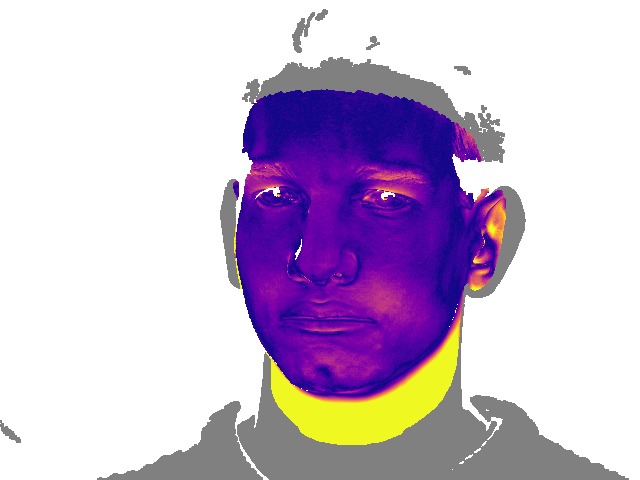} &
\includegraphics[width=23mm]{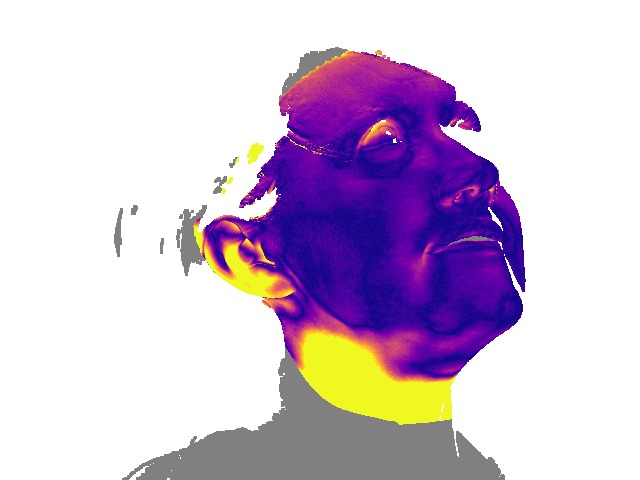} &
\includegraphics[width=23mm]{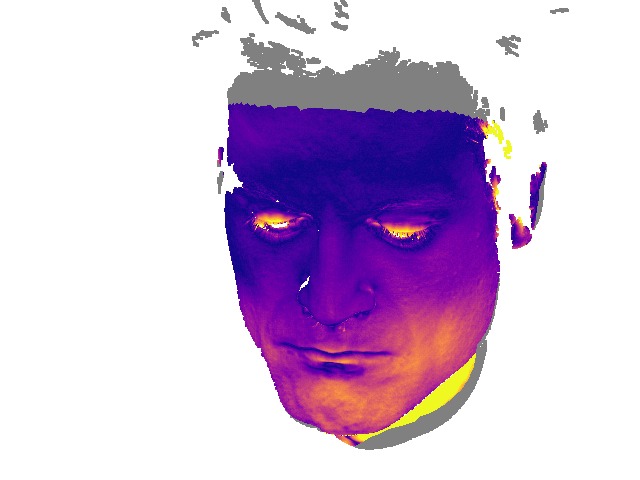} &
\includegraphics[width=23mm]{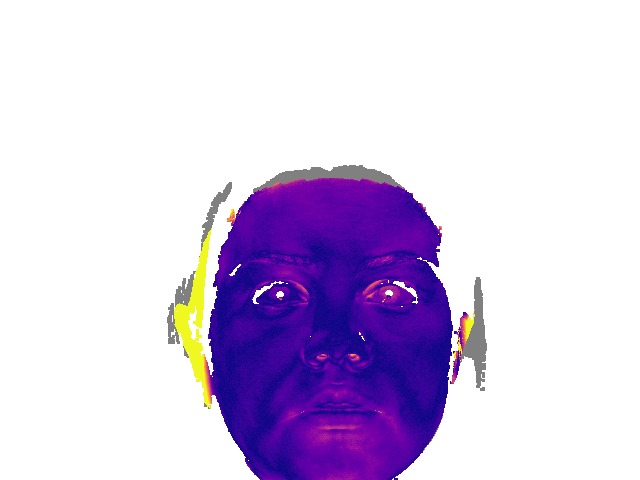} \\

Final Error &
&
\includegraphics[width=18mm]{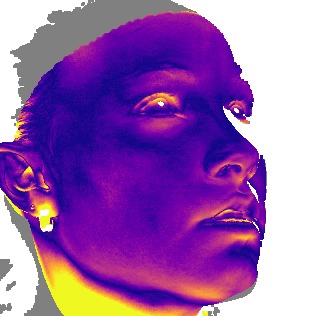} &
\includegraphics[width=18mm]{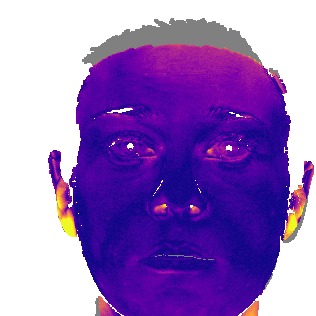} &
\includegraphics[width=18mm]{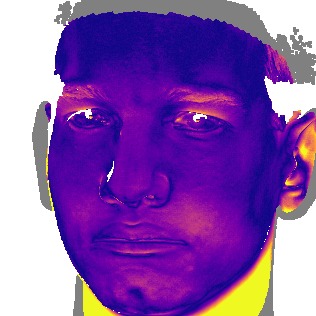} &
\includegraphics[width=18mm]{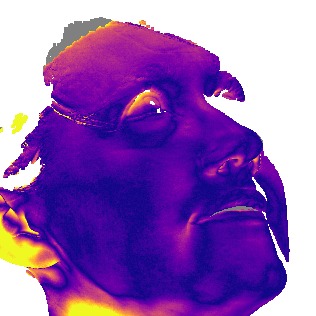} &
\includegraphics[width=18mm]{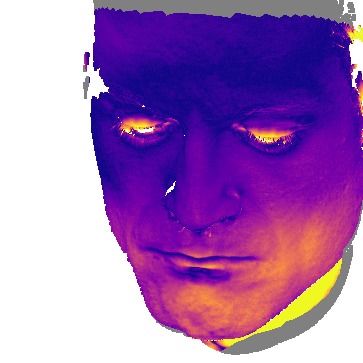} &
\includegraphics[width=18mm]{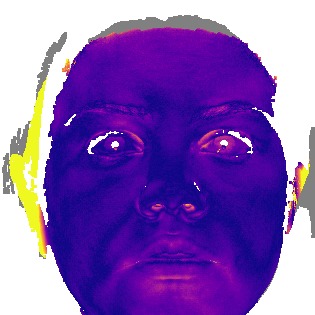} \\

Final Mesh&
 &
\includegraphics[width=18mm]{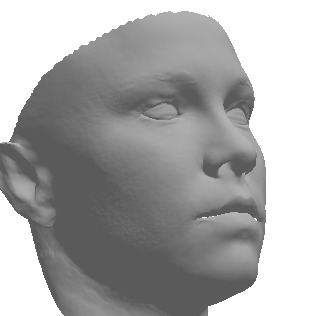} &
\includegraphics[width=18mm]{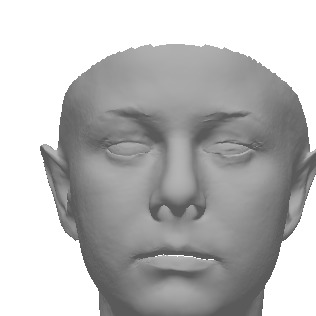} &
\includegraphics[width=18mm]{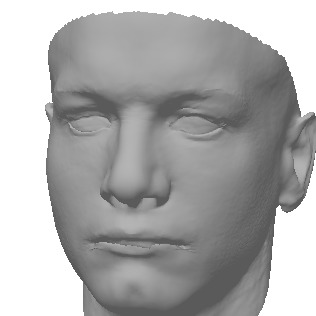} &
\includegraphics[width=18mm]{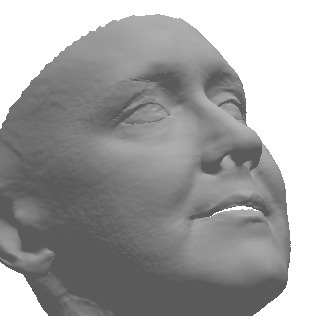} &
\includegraphics[width=18mm]{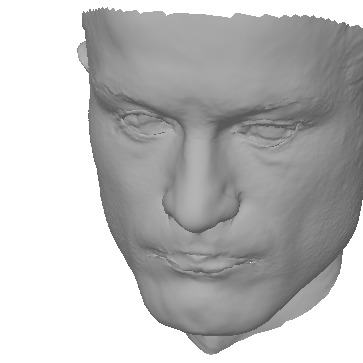} &
\includegraphics[width=18mm]{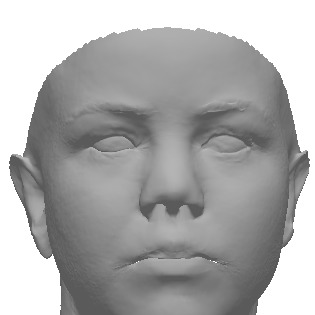} \\

Final Mesh Close Up&
&
\includegraphics[scale=0.36]{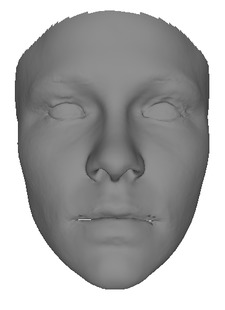} &
\includegraphics[scale=0.36]{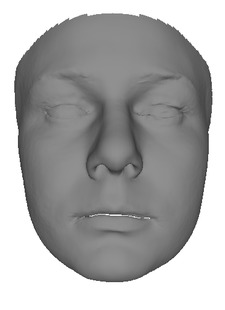} &
\includegraphics[scale=0.36]{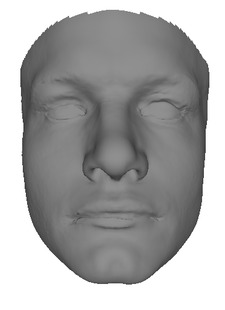} &
\includegraphics[scale=0.36]{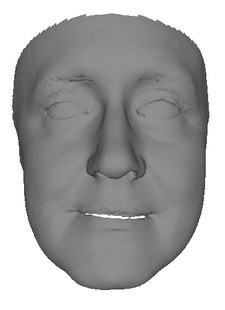} &
\includegraphics[scale=0.36]{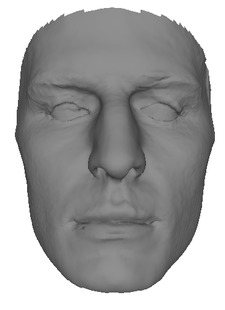} &
\includegraphics[scale=0.36]{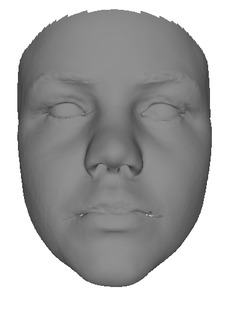} \\

\end{tabular}
\centering
\caption{For each image, the error image is shown for the initial alignment error and final alignment error. Error is shown at the scale of the image and also zoomed around the face for easier viewing. Additionally an initial and final mesh for each subject are shown in the same pose as the scan. Finally, a tight crop around the face is shown for all of the initial and final meshes.}
\label{table:big_table}
\end{figure*}

Some faces at the first and last iteration are shown along with the magnitude of the mean offsets in Figure~\ref{table:big_table}. The magnitudes of the offsets decrease as the error goes down.

\section{Future Work}
The method presented in this paper was used on facial scans collected during several scanning sessions. However the method's invariance to pose, invariance to outliers, and synthesis of multiple scans could make it especially useful in processing 3D video data.

Future research which follows this iterative paradigm of improving alignment and mesh geometry could use different methods for multi scan alignment or mesh warping. New methods could improve accuracy or reduce the total number of iterations required.

\section{Conclusion}\label{Conclusion}
The presented method addresses the problems of template warping and scan alignment in a joint framework. After steps are taken to process the scans, an iterative algorithm operates by first aligning scans to their subject's mesh. The new aligned scans update the mesh, which in turn is used to update the scan alignment.

Scans are aligned to the final subject mesh with mean distance less than one half millimeter. Furthermore, our method makes use of multiple scans taken in a wide variety of poses, while being robust to mislabeled data. We show that our iterative method for error reduction is well behaved with the mean error decreasing at each iteration. We experiment with a new detail parameter and show how this parameter can be used to further improve results.




\section*{ACKNOWLEDGMENT}
This research is based upon work supported by the Office of the Director of National Intelligence (ODNI), Intelligence Advanced Research Projects Activity (IARPA) under contract number 2014-14071600010. The views and conclusions contained herein are those of the authors and should not be interpreted as necessarily representing the official policies or endorsements, either expressed or implied, of ODNI, IARPA, or the U.S. Government. The U.S. Government is authorized to reproduce and distribute reprints for Governmental purpose notwithstanding any copyright annotation thereon.

\bibliography{bib}{}
\bibliographystyle{plain}
\end{document}